\documentclass{article}

\usepackage{microtype}
\usepackage{graphicx}
\usepackage{subfigure}
\usepackage{booktabs} % for professional tables

\usepackage{amsmath}
\usepackage{amsfonts}
\usepackage{amssymb}

\usepackage{hyperref,amsfonts}

\usepackage{tikz}
\usepackage{environ}

\usepackage[accepted]{icml2019_phys4dl}

\icmltitlerunning{Covariance in Physics and Convolutional Neural Networks}

\begin{document}
\usetikzlibrary{arrows.meta}

\twocolumn[
\icmltitle{Covariance in Physics and Convolutional Neural Networks}




\begin{icmlauthorlist}
\icmlauthor{Miranda C. N. Cheng}{kdv,iop,cnrs}
\icmlauthor{Vassilis Anagiannis}{iop}
\icmlauthor{Maurice Weiler}{quva}
\icmlauthor{Pim de Haan}{qc,quva}
\icmlauthor{Taco S. Cohen}{qc}
\icmlauthor{Max Welling}{qc}
\end{icmlauthorlist}

\icmlaffiliation{iop}{Institute of Physics, University of Amsterdam, Amsterdam, the Netherlands}
\icmlaffiliation{kdv}{Korteweg-de Vries Institute for Mathematics, University of Amsterdam, Amsterdam, the Netherlands}
\icmlaffiliation{quva}{Qualcomm-University of Amsterdam (QUVA) Lab, Amsterdam, the Netherlands}
\icmlaffiliation{qc}{Qualcomm AI Research, Amsterdam, the Netherlands}
\icmlaffiliation{cnrs}{On leave from CNRS, France}

\icmlcorrespondingauthor{Vassilis Anagiannis}{V.Anagiannis@uva.nl} 
\icmlkeywords{Machine Learning, ICML}

\vskip 0.3in
]



\printAffiliationsAndNotice{\icmlEqualContribution} % otherwise use the standard text.

\begin{abstract}
In this proceeding we give an overview of the idea of covariance (or equivariance) featured in the recent development of convolutional neural networks (CNNs). We study the similarities and differences between the use of covariance in theoretical physics and in the CNN context. Additionally, we demonstrate that the simple assumption of covariance, together with the required properties of locality, linearity and weight sharing, is sufficient to uniquely determine the form of the convolution. 
\end{abstract}

\section{Covariance and Uniqueness}

It is well-known that the principle of covariance, or coordinate independence, lies at the heart of the theory of  relativity. 
The theory of special relativity was constructed to describe Maxwell's theory of electromagnetism in a way that satisfies the special principle of covariance, which states ``If a system of coordinates $K$ is chosen so that, in relation to it, physical laws hold good in their simplest form, the same laws hold good in relation to any other system of coordinates $K'$ moving in uniform translation relatively to $K$'' \cite{einstein1916foundation}.

The transformation between $K$ and $K'$, in other words between different inertial frames, can always be achieved through an element of  the (global) Lorentz group. 
With the benefit of hindsight, there is no good reason why  physics should only be covariant under a global change of coordinates. 
Indeed, soon after the development of special relativity, Einstein started to develop his ideas for a theory that is covariant with respect to a local, spacetime-dependent, change of coordinates. In his words, the general principle of covariance states that ``The general laws of nature are to be expressed by equations which hold good for all systems of coordinates, that is, are covariant with respect to any substitutions whatever (generally covariant)'' \cite{einstein1916foundation}. The rest is history: the incorporation of the mathematics of Riemannian geometry in order to achieve general covariance and the formulation of the general relativity (GR) theory of gravity. It is important to note that the seemingly innocent assumption of general covariance is in fact so powerful that it determines GR as the unique theory of gravity compatible with this principle, and the equivalence principle in particular, up to short-distance corrections\footnote{The uniqueness argument roughly goes as follows. In order to achieve full general covariance,  the only ingredients for the gravitational part of the action are the Riemann tensor and its derivatives, with all the indices contracted. From a simple scaling argument one can show that all of them apart from the Ricci scalar have subleading effects at large distances. Note that this universality makes no assumptions on the matter fields that may be coupled to gravity.}.

In a completely different context, it has become clear in recent years that a coordinate-independent description is also  desirable for convolutional networks.
A covariant inference process is particularly useful in situations where the distribution of characteristic patterns is symmetric.
Important practical examples include satellite imagery or biomedical microscopy imagery which often do not exhibit a preferred global rotation or chirality. % {\bf helicity/chirality ???}.
In order to ensure that the inferred information of a network is equivalent for transformed samples, the network architecture has to be designed to be \emph{equivariant}\footnote{In this article we will use the words ``equivariant" and ``covariant" interchangeably as they convey the same concept.} under the corresponding group action\footnote{
To define the equivariance of a map requires the definition of a group action on the domain and codomain. One specific choice of group action is the co/contravariant transformation of tensors.}.
A wide range of equivariant models has been proposed for signals on flat Euclidean spaces $\mathbb{R}^d$.
In particular, equivariance w.r.t. (subgroups of) the Euclidean groups $\operatorname{E}(d)$ of translations, rotations and mirrorings of $\mathbb{R}^d$ has been investigated for planar images ($d=2$) \cite{cohenGroupEquivariantConvolutional2016, cohenSteerableCNNs2017, worrallHarmonicNetworksDeep2017, weilerLearningSteerableFilters2018, hoogeboomHexaConv2018}.
and volumetric signals ($d=3$)
\cite{winkels3DGCNNsPulmonary2018, worrallCubeNetEquivariance3D2018, weiler3DSteerableCNNs2018}
and has generally been found to outperform non-equivariant models in accuracy and data efficiency.
Equivariance has further proven to be a powerful principle in generalizing convolutional networks to more general spaces like the sphere \cite{cohenSphericalCNNs2018}.
In general, it has been shown in 
\cite{kondor2018generalization,cohenIntertwinersInducedRepresentations2018, cohenGeneralTheoryEquivariant2018}
that (globally) $H$-equivariant networks can be generalized to arbitrary homogeneous spaces $H/G$ where $G\leq H$ is a subgroup\footnote{Note that we are using an inverted definition of $H$ and $G$ w.r.t. the original paper to stay compatible with the convention used in \cite{cohen2019gauge}, discussed below.}.
The feature spaces of such networks are formalized as spaces of sections of vector bundles over $H/G$, associated to the principal bundle $H\to H/G$.
Our previous examples are in this setting interpreted as $\operatorname{E}(d)$-equivariant networks on Euclidean space $\mathbb{R}^d=\operatorname{E}(d)/\operatorname{O}(d)$ and $\operatorname{SO}(3)$-equivariant networks on the sphere $S^2=\operatorname{SO}(3)/\operatorname{SO}(2)$.
This description includes Poincar\'e-equivariant networks on Minkowski spacetime since Minkowski space $\mathbb{R}^{1,3}$ arises as quotient of the Poincar\'e group $\mathbb{R}^{1,3}\rtimes\operatorname{O}(1,3)$ w.r.t. the Lorentz group $\operatorname{O}(1,3)$.

Note that the change of coordinates required here is a global one.
Global symmetries are extremely natural and readily applicable when the underlying space is homogeneous, i.e. the group action is transitive, meaning the space contains only a single orbit.
At the same time, it is clearly desirable to have an effective CNN on an arbitrary surface, often not equipped with a global symmetry.
If the previous work on homogeneous spaces is based on an equivariance requirement analogous to the special principle of covariance, then what one needs for general surfaces is an analogue of  the general principle of covariance. In other words, we would like to have covariance with respect to a local, location-dependent coordinate transformations.

This requirement for local transformation equivariance of convolutional networks on general manifolds has been recognized and described in \cite{cohen2019gauge}.
A choice of local coordinates is thereby formalized as a \emph{gauge} $w_x:\mathbb{R}^d\to T_xM$ of the tangent space\footnote{
    The gauge is equivalent to choosing a basis $\{\tilde{e}_i\}_{i=1}^d:=\{w_x(e_i)\}_{i=1}^d$ of the tangent space by mapping the standard basis $\{e_i\}_{i=1}^d$ of $\mathbb{R}^d$ to $T_xM$.
    Explicitly, a coefficient vector $(v_1,\dots,v_d)$ determines a vector $w_x((v_1,\dots,v_d))=\sum_i v_i\tilde{e}_i$.
    Coordinate bases and frame (vielbein) bases are examples of choices of the gauge.
}.
Similar to the general theory of equivariant networks on homogeneous spaces, the feature fields of these networks are realized as sections of vector bundles over $M$, this time associated to the frame bundle $FM$ of the manifold. 
% in the case that $w_x$ is chosen to correspond to an orthonormal basis. 
Local transformations are described as position-dependent gauge transformations $w_x\mapsto w_x g_x$, where $g_x\in G\leq\operatorname{GL}(\mathbb{R}^d)$ is an element of the structure group. When the frame bundle is chosen to be the orthonormal frame bundle, the structure group is reduced to $O(d)$ and in our analogy this corresponds to the 
vierbein formulation of GR where the group $G$ is  the Lorentz group $\operatorname{O}(1,3)$. 

%{\bf MC: I propose to delete this paragraph, or to cut and move it. Since the story line is from global to local, I find it confusing to return to global and repeat the paragraph in the beginning of this page. It also severes the tie to the paragraph "The parallel....". But maybe I'm missing something.}
%Similar to a Lorentz covariant description implying global Poincar\'e covariance if the underlying space is Minkowskian \textbf{and locality is assumed ???},
%gauge equivariant networks can be shown to be equivariant under global isometries of the underlying manifold.
%Specifically, if the manifold is homogeneous, i.e. $M=H/G$ with structure group $G$, gauge equivariance implies a global equivariance under the action of $H$.

Note that the parallel between the two problems regarding general covariance forces us to employ the same mathematical language of (psuedo) Riemannian geometry. Interestingly, we will argue in the next section that our formalism is basically unique once general covariance along with some basic assumptions is demanded.  This can be compared with the long-distance uniqueness of GR, once covariance is required.

\section{The Covariant Convolution}

In CNNs we are interested in devising a linear map between the input feature space and the output space between every pair of subsequent layers in the convolutional network.
In this section we will argue that the four properties of 1) {\it linearity}, 2) {\it locality}, 3) {\it covariance}, and 4) {\it weight sharing}
is sufficient to uniquely determine its form, which we give in (\ref{eqn:int}) and (\ref{eqn:share}) below.

In mathematical terms, we describe the feature space in the $i$-th layer in terms of a fiber bundle $E_{\rm\it i}$ with fiber $F_{\rm\it i}$, that is associated to the principal bundle $P$ with the representation $\rho$ of the structure group $G$, with the projection $\pi : E_{\rm\it i} \to M$. The bundle structure  captures the transformation properties of the feature fields under a change of coordinates of the manifold $M$. For now we focus on a sub-region $U$ of the surface  which admits a  single coordinate chart and a local trivialisation of the bundles.
In this language, the feature field corresponds to a local section\footnote{Without the risk of causing confusion we will sometimes consider $f_i$ as a map from the local region $U$ to the fiber $F_i$, where implicitly we have used the local trivialisation to write $f_i(x)=(x,\alpha)$ where $\alpha\in F_i$ and ignored the first entry. } $f_{\rm\it i} \in \Gamma_i := \Gamma(E_{\rm\it i},U)$ of the fiber bundle and the linear map is between the space of sections: $m\in {\rm Hom}(\Gamma_{\rm \it in},\Gamma_{\rm \it out})$. Moreover, we require the linear map to satisfy the following {\it locality} condition: given the distance function $\|~\|$  on the manifold $M$, which in our case will be supplied by the metric, we have 
\(
(m\circ f)(x) = (m\circ \tilde f)(x) 
\)
for all $f, \tilde f \in \Gamma_{\rm \it in}$ with the property that $f(y)=\tilde f(y)$ for all $y$ with $\|y,x\|<R$ for  some (fixed) positive number $R$.
The linearity and the locality of the map immediately leads to the following form of the map. To illustrate this, consider the simplified scenario when the in- and output features are just numbers (scalars) which do not transform under coordinate transformation and $M$ is replaced with a set $\cal S$ with finite elements equipped with the distance function, then the above requirements immediately leads to the matrix form of the map $(m\circ f)(x) =\sum_{{y\in {\cal S}}, \|y,x\|<R}c_{x,y}f(y)$. Similarly, for our case we are led to the linear map
\begin{equation}\label{eqn:1}
f_{\rm \it out}(x)=(m\circ f_{\rm \it in})(x)  = \int_{b_{x,R}} k(x,y)\, f_{\rm \it in}(y)\, d^d y 
\end{equation}
where $d={\rm dim}M$,  $b_{x,R}$ is the ball centered at $x$ with radius $R$,  and $k : M\times M \to {\rm Hom}(F_{\rm \it in},F_{\rm \it out})$ is what will turn out to be the convolution kernel.

In the next step we will impose the condition of general covariance to restrict the form of $k(x,y)$. In the case of homogeneous spaces and when we require just special covariance, we can phrase the problem in the following general form. Suppose that the input feature and the output feature form representations $\rho_{\rm \it in}$ and $\rho_{\rm \it out}$ under a group $G$, then it is clear from the consistency of the $G$-action with the above map that $k$ must transform as $g: k \mapsto \rho_{\rm \it out}(g) k \rho_{\rm \it in}(g^{-1})$ and this is precisely what is described in \cite{cohenGeneralTheoryEquivariant2018,cohen2019gauge}. Once we promote the group element to be location-dependent, the analogous requirement is $ k(x,y) \mapsto \rho_{\rm \it out}(g_x) k(x,y) \rho_{\rm \it in}((g_y)^{-1})$. In our case, the group under discussion is that of local changes of coordinates\footnote{In this proceeding we mainly work with the basis $\{\partial_\mu\}$ of the tangent space, while another common choice is the orthonormal (vielbein) basis. The former has the advantage of being directly related to the covariance in the context of GR and the latter has the advantage of being closer to the philosophy of gauge theory.  }, with the consistent corresponding change of metric \(ds^2= g_{\mu\nu}(x)dx^\mu dx^\nu = g'_{\mu\nu}(x')(dx')^\mu (dx')^\nu\).  Note that this is not just a mathematical formality: one needs to deal with changes of coordinates when working with manifolds that cannot be covered with one coordinate chart, such as a sphere.

However, it is unwieldy to work with group elements at different points $x$ and $y$. Instead, we would like to encode the information in another way so that we can work with gauge/coordinate transformations at one single point when talking about the transformation of $k$.  
Here the relevant concept is  {\it parallel transport}. Given a bundle with connection $(E,\nabla)$ on $M$ and path $\gamma:I=[0,1]\to M$ with $\gamma(0)=y$ and $\gamma(1)=x$, for every $t_0\in  [0,1]$ and $s_0 \in E_{\gamma(t_0)}$ there is a unique section $s$ along $\gamma(I)\subset M$ that is flat along $\gamma$ such that $s(\gamma(t_0))=s_0$. In coordinates, this means ${dX^\mu\over dt} \nabla_\mu s(\gamma(t)) =0$ for all $t\in I.$ Note that the parallel transport is generically path-dependent; in other words, transporting $y$ to $x$ along different paths yields different results unless the bundle is flat. 

However, in our application we always have a uniquely distinguished path between $y$ and $x$ in practice. Namely, in the CNN context we let the ball $B_{x,R}$ containing the support of $k$ to be so small that every point in the ball is uniquely connected by a single geodesic to the center $x$. We hence replace $f_{\rm \it in}(y)$ with $f_{\rm \it in}\!\mid_y(x)$,  the parallel transport of $f_{\rm \it in}(y)$ along the unique geodesic from $y$ to the center point $x$. Denote the corresponding new kernel by $k'(x,y)$, and we arrive at the transformation property
\begin{equation}\label{eqn:Ktrans} k'(x,y) \mapsto \rho_{\rm \it out}(g_x) k'(x,y) \rho_{\rm \it in}((g_x)^{-1}). \end{equation} 
In fact, this geodesic description of the points provides us with an alternative, convenient way to parametrise the points we integrate over. Let $v\in T_xM$ be a vector in the tangent space at $x$.   
 There is a unique geodesic flow $\gamma_v:I=[0,1]\to M$ starting from $\gamma(0)=x$ where the initial velocity is $v$, i.e. ${dX^\mu(\gamma_v(t)) dt}\mid_{t=0}=v^\mu$. We will denote the endpoint of this flow ${\rm exp}_xv:=\gamma_v(1).$
  We can hence trade the integration within a small ball in our manifold with an integration within a ball of some radius $r$ in its tangent space ${B}_x:=\{v\in T_xM, v^\mu v^\nu g_{\mu\nu}(x)< r^2\}\subset T_xM$. We can hence write the kernel as $k''(x,v) \in {\rm Hom}(F_{\rm \it in},F_{\rm \it out})$.  
  
Apart from accommodating the transformation of the in- and out- feature fields, one also have to make sure that the integration measure remains invariant. From here we conclude that  should contain the factor of the volume form $\sqrt{|g(x)|} d^dv$, and we can hence write the kernel as $\sqrt{|g(x)|} k''(x,v)$. Note also that this volume factor $\sqrt{g}$ is simply 1 if one works with gauge $w_x$ that corresponds to an orthonormal basis.

At this stage, we arrive at the following form of our linear map 
 \begin{equation}\label{eqn:2}
f_{\rm \it out}(x)= \int_{B_{x}} \sqrt{|g(x)|} \, k''(x,v)\, f_{\rm \it in}\!\mid_{{\rm exp}_xv}(x)\, d^d v 
\end{equation}
From this stage onwards, we would like to be less abstract and focus on the groups and representations we encounter in real problems. Namely, when $E_i$ is the tensor product of the tangent and the cotangent bundles and takes the form $TM^{\otimes n}\otimes {T^\ast M}^{\otimes m}$ for $n,m\geq 0$. To see that this is sufficient and to make contact with previous work, note for instance  that any irreducible representations of $SO(3)$, denoted by ${\bf j}$, the spin $j$ irreducible representation with dimensions $2j+1$, can be expressed as an $\mathbb Z$-linear combination of the tensor product of the vector representation $V$. 
Specifically, we have $V^{\otimes n}={\bf n} \oplus \dots$, where $\dots$ denotes the direct sum of non-negative copies of ${\bf m}$ with  $0\leq m <n $,  for all $n\in \mathbb Z_{\geq 0}$. In other words, $V^{\otimes n}$ contains precisely one copy of spin-$n$ irreducible representations as well as other irreducibles with lower spins.
This ensures that every ${\bf j}$ is in the $\mathbb Z$-span of 1, $V, V^{\otimes 2}, \dots, V^{\otimes j}$. In other words, $\{{\bf j}, j=0,1,2,\dots\}$ as well as $\{V^{\otimes n}, n=0,1,2,\dots\}$ are equally good $\mathbb Z$-bases for $SO(3)$ representations. 

To ease notation, we will assume that the input and output feature fields are sections of tensor products of tangent bundles, while the cases involving the cotangent bundles can be treated with a straightforward generalisation of our formula. In this case, we can write explicit expressions for the output feature field as $ f_{\rm \it out}^{\mu_1\mu_2\dots \mu_{N_o}}$ and similarly for the input, and the  transformation property (\ref{eqn:Ktrans}) is succinctly summarised by the tensor and index structure of the kernel function, which we write as $K^{\mu_1\mu_2\dots \mu_{N_o}}_{\nu_1\dots \nu_{N_i}}(x,v)$.
In other words for a fixed $x$ and  $v\in T_xM$, we have $K(x,v) \in (T_xM)^{\otimes N_o} \otimes  (T_x^\ast M)^{\otimes N_i}$. 
Explicitly, we now have for these cases %{\bf MC: fix the typesetting}
% \begin{equation}\label{eqn:int}\scalebox{0.7}{ 
% $f_{\rm \it out}^{\mu_1\mu_2\dots \mu_{N_o}}(x)= \int_{B_{x}} \sqrt{|g(x)|} \, K^{\mu_1\mu_2\dots \mu_{N_o}}_{\nu_1\dots \nu_{N_i}}(x,v)\, f^{\nu_1\dots \nu_{N_i}}_{\rm \it in}\!\mid_{{\rm exp}_xv}(x)\, d^d v 
%$}
%\end{equation} 
\begin{align}\label{eqn:int}
&f_{\rm \it out}^{\mu_1\mu_2\dots \mu_{N_o}}(x)=\\
&\int_{B_{x}} \sqrt{|g(x)|} \, K^{\mu_1\mu_2\dots \mu_{N_o}}_{\nu_1\dots \nu_{N_i}}(x,v)\, f^{\nu_1\dots \nu_{N_i}}_{\rm \it in}\!\mid_{{\rm exp}_xv}(x)\, d^d v 
\nonumber
\end{align}

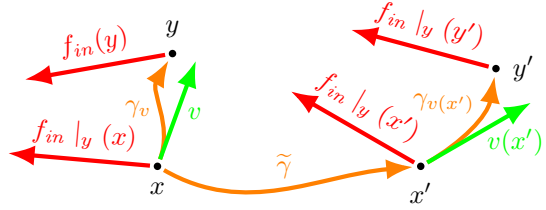
\begin{figure}
    \centering
    \begin{tikzpicture}[ultra thick]

\coordinate (X) at (0.0,0.0);
\coordinate (Xp) at (3.5,0.0);
\coordinate (Y) at (0.2,1.5);
\coordinate (Yp) at (4.5,1.3);

\node [fill=black, circle, inner sep=1, label=below:$x$] at (X) {};
\node [fill=black, circle, inner sep=1, label=above:$y$] at (Y) {};
\node [fill=black, circle, inner sep=1, label=below:$x'$] at (Xp) {};
\node [fill=black, circle, inner sep=1, label=right:$y'$] at (Yp) {};

% X -> Y
\draw [-Latex, shorten <= 0.1cm, shorten >= 0.1cm, orange] (X) to[out=70,in=-120] node[midway,left] {$\gamma_v$} (Y);

% X' -> Y'
\draw [-Latex, shorten <= 0.1cm, shorten >= 0.1cm, orange] (Xp) to[out=30,in=-100] node[midway,left,yshift=0.3cm,xshift=0.2cm] {$\gamma_{v(x')}$} (Yp);

% v(x)
\draw [-Latex, shorten <= 0.1cm, green] (X) -- node[midway,right] {$v$} ++(70:1.5cm);

% v(x')
\draw [-Latex, shorten <= 0.1cm, green] (Xp) -- node[midway,right,yshift=-0.1cm] {$v(x')$} ++(30:1.7cm);

% X -> X'
\draw [-Latex, shorten <= 0.1cm, shorten >= 0.1cm, orange] (X) to[out=-30,in=-175] node[midway,above] {$\widetilde\gamma$} (Xp);

% f(y)
\draw [-Latex, shorten <= 0.1cm, red] (Y) -- node[midway,sloped,above] {$f_{in}(y)$} ++(-170:2cm);
\draw [-Latex, shorten <= 0.1cm, red] (X) -- node[midway,sloped,above] {$f_{\rm \it in}\mid_{y}(x)$} ++(175:2.0cm);
\draw [-Latex, shorten <= 0.1cm, red] (Xp) -- node[midway,sloped,above] {$f_{\rm \it in}\mid_{y}(x')$} ++(150:2.0cm);
\draw [-Latex, shorten <= 0.1cm, red] (Yp) -- node[midway,sloped,above] {$f_{\rm \it in}\mid_{y}(y')$} ++(165:2.0cm);

\end{tikzpicture}
    \caption{Parallel transport of the feature field.}
    \label{fig:transport}
\end{figure}

 Finally, we would like to impose the weight sharing condition, which we phrase in the following way: {\it when the (localised) input signal is parallel transported along a curve, the output signal should also equal to the parallel transport of the previous result}. First, we need to explain what we mean by parallel transporting the input feature field along a curve $\tilde \gamma : [0,1] \to M$ with $\tilde \gamma (0) = x$ and $\tilde \gamma (1) = x'$. For the point $x$ itself it is clear that we can simply parallel transport $f_{\rm \it in}\mid_{x}(x)$ to $f_{\rm \it in}\mid_{\tilde \gamma(t)}(\tilde \gamma(t))$.  Suppose that $y={\rm exp}_xv$ is connected by the geodesic flow $\gamma_v$ starting from $x$, we also transport $v\in T_xM$ to $v(\tilde \gamma (t)) \in T_{\tilde \gamma (t)}M$ and  transport
 $f_{\rm \it in}\!\mid_{y}(x)$ by  transporting it to  $f_{\rm \it in}\mid_{y}(\tilde \gamma(t))$ and then further transport it along the geodesic flow $\gamma_{v(\tilde \gamma (t) )}$ starting from $\tilde \gamma (t)$. After this prescription, depicted in Figure~\ref{fig:transport}, it is clear how one should define $K(x',v)$ such that the weight sharing condition is true. Recall that for a given $v\in T_xM$,  $K_v(x):=K(x,v) \in (T_xM)^{\otimes N_o} \otimes  (T_x^\ast M)^{\otimes N_i}$. Now parallel transport it along $\tilde \gamma$  to obtain $K_v(\tilde \gamma (t))$, and define that \begin{equation}\label{eqn:share}
 K(\tilde \gamma(t),v(\tilde \gamma (t) ) ) := K_v(\tilde \gamma (t)).\end{equation} In other words, we simultaneously parallel transport the dependence on the tangent vector. The vanishing of  the covariant derivative of the output feature along the curve, $\tilde \gamma^{\ast}(\nabla)f_{\rm \it out}(\tilde \gamma(t))$ then just comes from the vanishing of the   covariant derivative of volume form and the definition  (\ref{eqn:share}).  

Note that (\ref{eqn:share}) means that, along the path, the kernel $K(\gamma(t),v)$ is completely determined by the kernel $K(\gamma(t_0),v)$ at any point $t_0$ on the path. 
Supposed further that we select a reference point $x_\ast \in M$ on the manifold. For any point  
$y\in M$ that is connected to $x_\ast$ by a unique geodesic, parallel transport with respect to the geodesic then unambiguously ``share"  the kernel at $x_\ast$ with $y$. On the other hand, when $y$ is connected by more than one geodesics, the general covariance then dictates the relation between the outputs corresponding to different geodesics. Moreover, this covariance also holds for transporting along different paths (not necessarily geodesics) in general. More precisely, we see how different kernels, related again by a local change of coordinates, can be compensated by a transformation of the input and output feature fields. 
We hence see that our simple and general assumptions in fact completely determine the form of the convolution map.  

\section{Discussion}
After pointing out the parallel between special and general relativity and equivariant CNNs, it is also important to point out the crucial differences. 
In the CNN setup, the geometry is always held fixed  and we do not consider dynamics of the metric.  
From this point of view the closer analogy is perhaps the study of  field theories in a fixed curved spacetime where the back-reaction of the matter fields to the spacetime geometry has been ignored. It would be interesting to explore equivariant CNNs with geometry that evolves between layers in future work. It is certainly tempting to treat the direction of different layers as a part of the spacetime, either as the temporal or the holographic direction \cite{tHooft:1993dmi}. This interpretation is particularly relevant if all feature spaces carry the same group representation. 

\section*{Acknowledgements}
The work of MC is supported by ERC starting grant \#640159 and NWO Vidi grant ERC starting grant H2020 ERC StG \#640159. The work of VA is supported by ERC starting grant  \#640159. 

%\textbf{Do not} include acknowledgements in the initial version of the paper submitted for blind review.


\bibliography{refs}

\begin{thebibliography}{15}
\providecommand{\natexlab}[1]{#1}
\providecommand{\url}[1]{\texttt{#1}}
\expandafter\ifx\csname urlstyle\endcsname\relax
  \providecommand{\doi}[1]{doi: #1}\else
  \providecommand{\doi}{doi: \begingroup \urlstyle{rm}\Url}\fi

\bibitem[Cohen et~al.(2018{\natexlab{a}})Cohen, Geiger, and
  Weiler]{cohenGeneralTheoryEquivariant2018}
Cohen, T., Geiger, M., and Weiler, M.
\newblock A {{General Theory}} of {{Equivariant CNNs}} on {{Homogeneous
  Spaces}}.
\newblock 2018{\natexlab{a}}.

\bibitem[Cohen \& Welling(2016)Cohen and
  Welling]{cohenGroupEquivariantConvolutional2016}
Cohen, T.~S. and Welling, M.
\newblock Group equivariant convolutional networks.
\newblock In \emph{ICML}, 2016.

\bibitem[Cohen \& Welling(2017)Cohen and Welling]{cohenSteerableCNNs2017}
Cohen, T.~S. and Welling, M.
\newblock Steerable {{CNNs}}.
\newblock In \emph{ICLR}, 2017.

\bibitem[Cohen et~al.(2018{\natexlab{b}})Cohen, Geiger, Koehler, and
  Welling]{cohenSphericalCNNs2018}
Cohen, T.~S., Geiger, M., Koehler, J., and Welling, M.
\newblock Spherical {{CNNs}}.
\newblock In \emph{ICLR}, 2018{\natexlab{b}}.

\bibitem[Cohen et~al.(2018{\natexlab{c}})Cohen, Geiger, and
  Weiler]{cohenIntertwinersInducedRepresentations2018}
Cohen, T.~S., Geiger, M., and Weiler, M.
\newblock Intertwiners between {{Induced Representations}} (with
  {{Applications}} to the {{Theory}} of {{Equivariant Neural Networks}}).
\newblock 2018{\natexlab{c}}.

\bibitem[Cohen et~al.(2019)Cohen, Weiler, Kicanaoglu, and
  Welling]{cohen2019gauge}
Cohen, T.~S., Weiler, M., Kicanaoglu, B., and Welling, M.
\newblock Gauge equivariant convolutional networks and the icosahedral cnn.
\newblock \emph{arXiv preprint arXiv:1902.04615}, 2019.

\bibitem[Einstein(1916)]{einstein1916foundation}
Einstein, A.
\newblock The foundation of the general theory of relativity.
\newblock \emph{Annalen der Physik}, 49\penalty0 (7):\penalty0 769--822, 1916.

\bibitem[Hoogeboom et~al.(2018)Hoogeboom, Peters, Cohen, and
  Welling]{hoogeboomHexaConv2018}
Hoogeboom, E., Peters, J. W.~T., Cohen, T.~S., and Welling, M.
\newblock {{HexaConv}}.
\newblock In \emph{ICLR}, 2018.

\bibitem[Kondor \& Trivedi(2018)Kondor and Trivedi]{kondor2018generalization}
Kondor, R. and Trivedi, S.
\newblock On the generalization of equivariance and convolution in neural
  networks to the action of compact groups.
\newblock \emph{arXiv preprint arXiv:1802.03690}, 2018.

\bibitem['t~Hooft(1993)]{tHooft:1993dmi}
't~Hooft, G.
\newblock {Dimensional reduction in quantum gravity}.
\newblock \emph{Conf. Proc.}, C930308:\penalty0 284--296, 1993.

\bibitem[Weiler et~al.(2018{\natexlab{a}})Weiler, Geiger, Welling, Boomsma, and
  Cohen]{weiler3DSteerableCNNs2018}
Weiler, M., Geiger, M., Welling, M., Boomsma, W., and Cohen, T.~S.
\newblock {3D Steerable CNNs: Learning Rotationally Equivariant Features in
  Volumetric Data}.
\newblock In \emph{NIPS}, 2018{\natexlab{a}}.

\bibitem[Weiler et~al.(2018{\natexlab{b}})Weiler, Hamprecht, and
  Storath]{weilerLearningSteerableFilters2018}
Weiler, M., Hamprecht, F.~A., and Storath, M.
\newblock Learning {{Steerable Filters}} for {{Rotation Equivariant CNNs}}.
\newblock In \emph{CVPR}, 2018{\natexlab{b}}.

\bibitem[Winkels \& Cohen(2018)Winkels and Cohen]{winkels3DGCNNsPulmonary2018}
Winkels, M. and Cohen, T.~S.
\newblock {{3D G}}-{{CNNs}} for {{Pulmonary Nodule Detection}}.
\newblock In \emph{International {{Conference}} on {{Medical Imaging}} with
  {{Deep Learning}} ({{MIDL}})}, 2018.

\bibitem[Worrall \& Brostow(2018)Worrall and
  Brostow]{worrallCubeNetEquivariance3D2018}
Worrall, D.~E. and Brostow, G.~J.
\newblock Cubenet: Equivariance to 3d rotation and translation.
\newblock In \emph{ECCV}, 2018.

\bibitem[Worrall et~al.(2017)Worrall, Garbin, Turmukhambetov, and
  Brostow]{worrallHarmonicNetworksDeep2017}
Worrall, D.~E., Garbin, S.~J., Turmukhambetov, D., and Brostow, G.~J.
\newblock Harmonic {{Networks}}: {{Deep Translation}} and {{Rotation
  Equivariance}}.
\newblock In \emph{CVPR}, 2017.

\end{thebibliography}
\bibliographystyle{icml2019}

% \appendix
% \section{Acceptable to have an appendix here}

\end{document}